\documentclass[10pt,twocolumn,letterpaper]{article}

\usepackage{cvpr}              

\newcommand{\ours}[1]{\text{OPT-Pose}}
\usepackage{microtype}
\usepackage{algorithmic}
\usepackage{amssymb}
\usepackage{pifont}
\usepackage{mathtools}

\definecolor{cvprblue}{rgb}{0.21,0.49,0.74}
\usepackage[pagebackref,breaklinks,colorlinks,allcolors=cvprblue]{hyperref}
\usepackage{multirow}
\usepackage{placeins}
\usepackage{arydshln} 
\usepackage{booktabs}
\usepackage{enumitem}
\usepackage[table]{xcolor}
\usepackage{algorithm}
\usepackage{pifont}   

\def\paperID{} 
\def\confName{CVPR}
\def\confYear{2026}

\title{Object Pose Transformer: Unifying Unseen Object Pose Estimation}

\author{
Weihang Li\textsuperscript{1,2} \ \
Lorenzo Garattoni\textsuperscript{3} \ \
Fabien Despinoy\textsuperscript{3} \ \
Nassir Navab\textsuperscript{1,2} \\
Benjamin Busam\textsuperscript{1,2} \\
{\textsuperscript{1}Technical University of Munich} \quad
{\textsuperscript{2}Munich Center for Machine Learning} \quad
{\textsuperscript{3}Toyota Motor Europe}
}

\begin{document}

\newcommand{\warning}{\textcolor{red}}

\maketitle
{\let\thefootnote\relax\footnotetext{Project Page: \url{https://colin-de.github.io/OPT-Pose/}}}
\begin{abstract}
Learning model-free object pose estimation for unseen instances remains a fundamental challenge in 3D vision. Existing methods typically fall into two disjoint paradigms: category-level approaches predict absolute poses in a canonical space but rely on predefined taxonomies, while relative pose methods estimate cross-view transformations but cannot recover single-view absolute pose. In this work, we propose Object Pose Transformer (\ours{}), a unified feed-forward framework that bridges these paradigms through task factorization within a single model. \ours{} jointly predicts depth, point maps, camera parameters, and normalized object coordinates (NOCS) from RGB inputs, enabling both category-level absolute SA(3) pose and unseen-object relative SE(3) pose. Our approach leverages contrastive object-centric latent embeddings for canonicalization without requiring semantic labels at inference time, and uses point maps as a camera-space representation to enable multi-view relative geometric reasoning. Through cross-frame feature interaction and shared object embeddings, our model leverages relative geometric consistency across views to improve absolute pose estimation, reducing ambiguity in single-view predictions. Furthermore, \ours{} is camera-agnostic, learning camera intrinsics on-the-fly and supporting optional depth input for metric-scale recovery, while remaining fully functional in RGB-only settings. Extensive experiments on diverse benchmarks (NOCS, HouseCat6D, Omni6DPose, Toyota-Light) demonstrate state-of-the-art performance in both absolute and relative pose estimation tasks within a single unified architecture. 
\end{abstract} 

\begin{figure}[t] 
    \hspace*{\fill} 
    \begin{minipage}[t]{0.5\textwidth} 
        \centering
        \includegraphics[width=\textwidth]{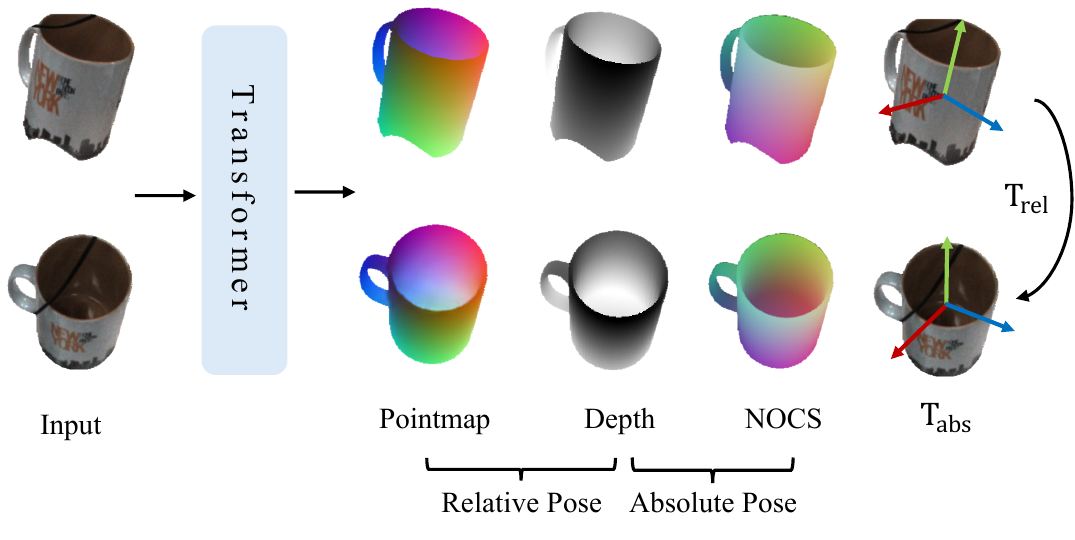} 
        \vspace{-0.2cm}
        \caption{\textbf{Unified unseen object pose estimation.}
        OPT-Pose utilizes a feed-forward transformer to predict point map, depth, NOCS, and camera parameters.
        Existing category-level methods predict canonical absolute 9-DoF SA(3) poses (equivalent to Depth + NOCS), but require predefined category labels and calibrated cameras. Relative pose methods align unseen objects across views in 6-DoF SE(3) (equivalent to Pointmap + Depth), but do not support single-view absolute pose prediction.
        OPT-Pose enables the simultaneous recovery of both unseen-object relative and category-level absolute poses (right-most column) for flexible single or multi-view RGB or RGB-D input, without the need for CAD models or semantic labels.
        }
        \vspace{-0.4cm}
        \label{fig:teaser}
    \end{minipage}
\end{figure}

\section{Introduction}
\label{sec:intr}
Object pose estimation for unseen object instances, without relying on prior CAD models, is fundamental in vision. It unlocks object understanding to enhance robotic manipulation, augmented reality, and autonomous systems. Existing model-free approaches follow two paradigms: \emph{category-level absolute pose estimation} predicts canonical-space 9-DoF SA(3) transforms for instances within known categories \cite{nocs,Li_2025_CVPR,chen2024secondpose,lin2024instance,fu2022category,tian2020shape,DualPoseNet,lin2023vi,liu2023istnet,i2c,Li_2025_ICCV} but relies on predefined taxonomies and category labels; \emph{relative pose estimation} aligns unseen objects across views via 6-DoF SE(3) \cite{labbe2020cosypose,labbe2022megapose,he2022oneposeplusplus,nguyen2024gigapose,ornek2025foundpose,corsetti2024open,corsetti2024high} but lacks canonicalization and cannot handle single-view absolute pose. However, both paradigms remain constrained. Category-level methods require explicit category names at inference \cite{chen2024secondpose,lin2023vi,DualPoseNet,huang_2025_GIVEPose,Li_2025_CVPR,liu2025diff9d,zhang2023genpose,zhang2024omni6dpose}, limiting their generalization to open-vocabulary conditions. Additionally, relative methods typically require multiple views and cannot handle single-view absolute pose. To the best of our knowledge, no prior work unifies these complementary tasks in a single category-agnostic model while leveraging their interplay to improve pose estimation and generalization to unseen objects.

We propose Object Pose Transformer (\ours{}), a unified feed-forward framework for \emph{model-free unseen object pose estimation with task factorization}. \ours{} unifies category-level absolute pose and unseen-object relative pose in a single model by predicting depth, point maps, camera parameters alongside NOCS from RGB images. The core design insight is a complementary geometric mechanism: 

\begin{itemize}
    \item \textbf{Canonical-space grounding.} Depth + NOCS align instances into a shared canonical space, enabling absolute SA(3) pose estimation without requiring category labels.
    \item \textbf{Relative geometric reasoning.} Depth + point maps represent objects in camera space, enabling multi-view SE(3) reasoning across frames that provides additional geometric constraints and improves absolute pose estimation.
\end{itemize}
This factorization bridges camera- and canonical-space reasoning without CAD models or predefined taxonomies.

\ours{} employs a multi-view transformer that aggregates image tokens and dispatches them to lightweight task heads. A keypoint-centric attention module builds soft correspondences over sampled pixels, while a visual--geometric fusion block integrates local 3D neighborhoods with global context to produce discriminative keypoint descriptors. These descriptors are pooled into an \emph{object latent embedding} and used to FiLM-condition the NOCS head~\cite{perez2018film}. We train this latent representation with a contrastive InfoNCE objective across views \cite{oord2018cpc}, enabling a shared canonical space without requiring semantic labels at inference time. Unlike methods that scale to hundreds of categories but still depend on predefined taxonomies, \ours{} is category-agnostic and treats all objects uniformly. A dedicated camera head estimates intrinsic parameters, enabling camera-agnostic operation. A metric-recovery head aggregates keypoint-level depth evidence when measured depth is available, enabling metric-scale recovery in RGB-D mode while remaining fully functional in RGB-only settings.

Extensive experiments across diverse datasets and tasks, including category-level absolute pose (REAL275, HouseCat6D, Omni6DPose~\cite{zhang2024omni6dpose,nocs,jung2024housecat6d}) and unseen-object relative pose (REAL275, Toyota-Light)\cite{hodan2018bop}, show that \ours{} achieves state-of-the-art performance on both absolute and relative pose estimation, generalizing across object categories, camera types, and input modalities (RGB/-D).

We summarize our contributions as follows:
\begin{itemize}
    \item \textbf{Unified Object Poses.}
    A unified model-free framework for category-level absolute SA(3) and unseen-object relative SE(3) pose via complementary geometric mechanisms. We leverage multi-view relative geometric reasoning to improve absolute pose estimation, without CAD model.
  \item \textbf{Category-Agnostic Canonicalization.}
  We learn a category-agnostic canonicalization for inference via a contrastive objective without class labels.
  \item \textbf{Flexible inputs and outputs.}
  \ours{} supports RGB and RGB-D inputs from single or multiple views, achieving state-of-the-art performance across both category-level absolute and relative unseen-object pose tasks.
\end{itemize}

\begin{figure*}[t]
    \centering
    \includegraphics[width=\textwidth]{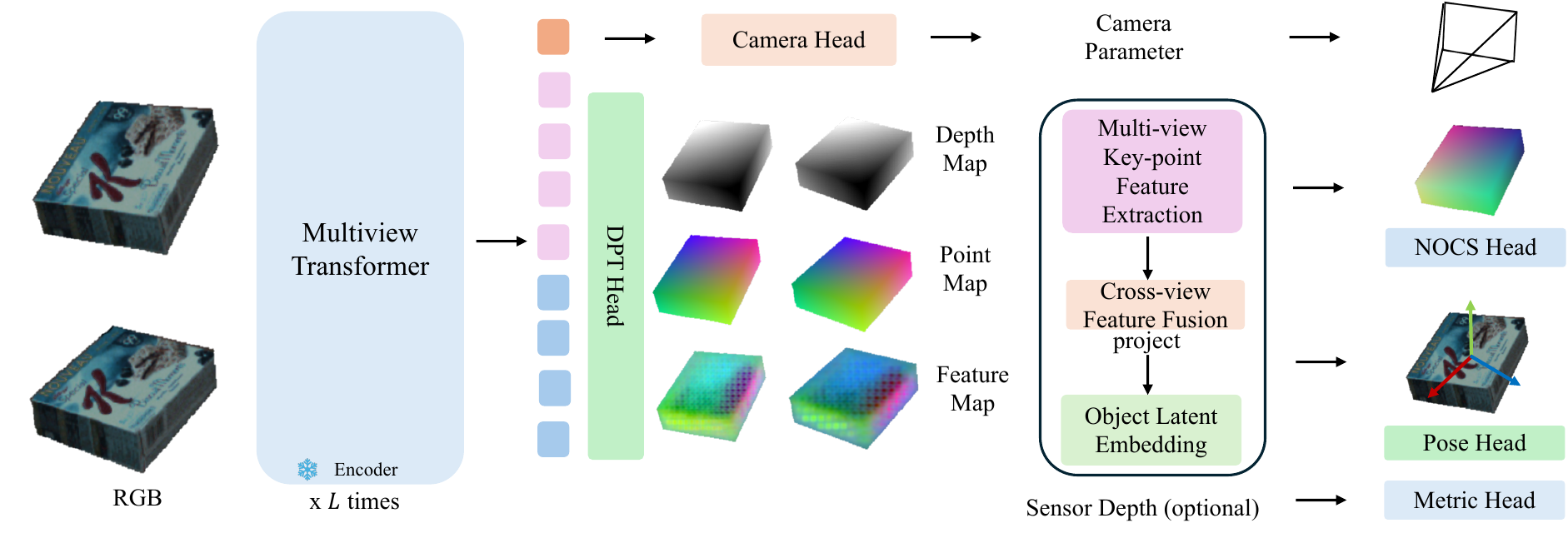}
    \vspace{-0.4cm}
    \caption{\ours{} overview. A multiview transformer aggregates image tokens and emits predictions from light heads: camera parameters, depth, and point maps for camera-space geometry; a multi-view-keypoint-centric module fuses RGB and 3D features to discover object keypoints, predict NOCS coordinates, and build an object latent embedding. Absolute pose (SA(3)) and relative pose (SE(3)) are recovered in a single forward pass. Optional sensor depth provides metric scale, while the system remains fully functional in RGB-only mode.}
    \vspace{-0.4cm}
    \label{fig:pipeline_overview}
\end{figure*}

\section{Related Works} \label{sec:related_works}

\paragraph{Category-level, Model-free Absolute Pose Estimation.}
Category-level methods lift objects into a canonical space using normalized object coordinate systems (NOCS)~\cite{nocs} and regress per-pixel correspondences or keypoints~\cite{lin2024instance,Li_2025_CVPR} to recover 9D poses. While recent category-level approaches focus on improving accuracy in pose prediction using shape or semantic priors \cite{chen2020learning,Chen_2021_CVPR,DualPoseNet,di2022gpv,Zheng_2023_CVPR,lin2024instance, zhang2024generative} or extending the existing category to large-vocabulary methods~\cite{zhang2024omni6dpose,krishnan2024omninocs,zhang2024omni6dlargevocabulary3dobject,ov9d}, they still require predefined taxonomies and explicit category names at inference, limiting true open-vocabulary deployment. Many existing methods with high accuracy, 
like GCE-Pose\cite{Li_2025_CVPR} and AG-Pose\cite{lin2024instance}, further design specialized networks with explicit categorical shape and semantic priors \cite{Li_2025_CVPR,lin2024instance,chen2024secondpose, lin2022category, wang2025gs}. These constraints prevent generic, category-agnostic operations. 

\paragraph{Feed-forward Geometry Transformers.}
Recent feed-forward vision transformers (e.g., DUSt3R~\cite{wang2024dust3r}, MASt3R~\cite{mast3r_eccv24}, CUT3R~\cite{cut3r}, VGGT~\cite{cut3r}) jointly predict geometric signals such as depth, camera parameters, and point maps in a single forward pass. They decouple representation learning from task-specific heads, enabling multi-view token aggregation and camera self-calibration. \ours{} extends this paradigm to object-centric perception by additionally learning keypoints, object-centric latent embeddings, and latent-conditioned NOCS, bridging camera- and canonical-space reasoning for joint absolute and relative pose estimation.

\paragraph{Relative Pose Estimation and Point Cloud Registration.}
Two-view relative pose estimation is commonly solved via feature matching and post optimization (e.g., essential matrix, PnP), or learned 2D/3D matching and registration \cite{epnp,Fischler1981RandomSC,he2022oneposeplusplus,labbe2022megapose,shugurov2022osop,wen2024foundationpose,nguyen2024gigapose}. For model-free methods, (H)Oryon~\cite{corsetti2024open,corsetti2024high} establish cross-view correspondences via feature matching and point cloud registration. Any6D and OnePoseViaGen~\cite{lee_any6d_2025,geng2025one} instead generate object meshes using image-to-3D diffusion~\cite{xu2024instantmesh,long2024wonder3d,wu2025amodal3r,xiang2025structured} and align them via render-and-compare. In contrast, we predict object-centric point maps and leverage measured depth to directly estimate SE(3) alignment with weighted Umeyama~\cite{umeyama1991least}.

\paragraph{Metric Scale Recovery.}
Monocular predictions are inherently scale-ambiguous. Previous work leveraged category-level size information and regressed offsets to real sizes \cite{zhang2024lapose,huang_2025_GIVEPose,fischer2025unified,msos,oldnet,dmsr}. In our setting, the optional sensor depth acts as an external signal. \ours{} predicts absolute translation/size from depth-derived points and derives the scale by comparing against normalized predictions. Additionally, in RGB-only mode, a lightweight head estimates per-frame log-scale.

\paragraph{Semantic Priors and Category-agnostic Canonicalization.}
Vision and language foundation models \cite{oquab2023dinov2,radford2021learning} provide rich semantic features, but existing category-level methods \cite{Li_2025_CVPR,lin2024instance,zhang2023genpose,chen2024secondpose} still require category labels at inference. In contrast, \ours{} projects object-centric latent embeddings, extracted from keypoint-level visual features~\cite{wang2025vggt} and geometric representations from Sonata~\cite{wu2025sonata}, into a shared object-latent space. We train this space using a contrastive objective across views, enabling category-agnostic canonicalization without requiring category names at inference.

\paragraph{Taxonomy of Unseen Pose Estimation and \ours{}.}
Model-free object pose estimation falls into two paradigms: (i) \emph{Category-level absolute pose estimation} (e.g., GCE-Pose, AG-Pose)~\cite{Li_2025_CVPR,lin2024instance}, which predicts SA(3) transforms in a canonical space but relies on predefined taxonomies; and (ii) \emph{Unseen-object relative pose estimation} (e.g., OnePose++, MegaPose, OSOP)~\cite{sun2022onepose,labbe2022megapose,shugurov2022osop}, which estimates SE(3) across views but cannot recover single-view absolute pose. \ours{} unifies these paradigms in a model-free framework via task factorization, jointly predicting canonical-space absolute pose and camera-space relative pose through complementary geometric mechanisms. This formulation removes the need for CAD models, category labels, and test-time optimization. In addition, \ours{} achieves category-agnostic generalization through contrastive latent learning, while remaining camera-agnostic and supporting flexible RGB(-D) inputs across single and multi-view settings.

\section{Method} \label{sec:method}

\subsection{Design and Task Factorization}
We address model-free, unseen-object pose estimation, covering category-level absolute and relative pose estimation between views, through a geometric observation: two complementary correspondence pairs are sufficient to link canonical and camera spaces.
\begin{enumerate}
  \item \textbf{Depth + NOCS} $\Rightarrow$ \textbf{category-level absolute} $\mathrm{SA}(3)$ \textbf{pose.} Depth provides metric 3D observations in camera space, whereas NOCS yields canonical correspondences; aligning these recovers rotation, translation, and scales for category-level reasoning.
  \item \textbf{Depth + Point Map} $\Rightarrow$ \textbf{unseen-object relative} $\mathrm{SE}(3)$ \textbf{pose.} The two 3D representations enable robust cross-frame alignment without explicit canonicalization.
\end{enumerate}
This factorization naturally supports single- and multi-view cases with RGB-(D), and is category- and camera-agnostic.

\subsection{Problem Formulation}
\label{subsec:problem}
Given a sequence of RGB frames $\{I_i\}_{i=1}^{S}$, we predict per-frame object geometry and poses, without CAD prior. Let $\mathbf{K}_i$ denote camera intrinsics, $\mathbf{T}_i\in \mathrm{SE}(3)$ the camera-to-world extrinsics (implicitly represented by our pose encoding), and $\mathbf{P}_i\in\mathbb{R}^{H\times W\times 3}$ the point map (i.e., the 3D camera-space coordinates). Let further $\mathcal{I}_i$ be a set of $K$ sampled pixels and $\mathbf{X}^{\text{obj}}_{i,k}\in\mathbb{R}^3$ their camera-space 3D points.

In a canonical object space (NOCS), object coordinates lie in $[-0.5,0.5]^3$. For each frame, we predict $M$ keypoints with canonical coordinates $\mathbf{C}_{i,m}\in\mathbb{R}^3$, and corresponding 3D observations $\mathbf{X}^{\text{obj}}_{i,m}\in\mathbb{R}^3$ in the first (anchor) camera space. The (absolute) category-level object pose can be retrieved by aligning the canonical to the observed coordinates via transformation in the rigid anisotropic similarity transformation

\begin{align}
\mathrm{SA}(3) &\coloneqq \mathbb{R}^3 \times SO(3) \times \mathrm{Diag}^+(3) \\
    \text{with}\quad \mathbf{X}^{\text{obj}}_{i,m} &= \mathbf{R}_i \mathbf{S} \mathbf{C}_{i,m} + \mathbf{t}_i
\end{align}

and $\mathbf{S} = \mathrm{diag}(s_i)_{i=1}^3$ with $\mathrm{SE}(3) \subset \mathrm{SA}(3)$ for $\mathbf{S} = \mathbf{I}$ and $\mathrm{Sim}(3) \subset \mathrm{SA}(3)$ for $\mathbf{S} = s\mathbf{I},\ s>0$.
We also estimate a relative pose $\Delta\mathbf{T}_i\in\mathrm{SE}(3)$ aligning the two geometry branches (i.e., depth-derived vs.~point-map predictions).

\subsection{Multiview Geometry and Feature Transformer}
\label{subsec:overview}
Our model is a single feed-forward multiview transformer that produces all geometric outputs jointly. (see. Fig.~\ref{fig:pipeline_overview}). The aggregator encodes each image into a sequence of tokens across several refinement iterations, using a visual backbone \cite{wang2025vggt, oquab2023dinov2} with frozen patch embedding for stable training. From these tokens, the camera head predicts per-frame intrinsics through a field-of-view representation and extrinsics as quaternions. The depth head estimates dense depth $\hat{D}_i$ and confidence, which we convert to camera-space points and normals; at sampled indices $\mathcal{I}_i$, we gather points $\mathbf{X}^{\text{obj}}_{i,k}$ together with colors and normals for keypoint reasoning. A parallel point-map head predicts a dense 3D point map $\hat{\mathbf{P}}_i$, and we extract $\mathbf{X}^{\mathrm{pm}}_{i,k}$ as an independent structural cue. A canonicalization head predicts keypoint-level NOCS coordinates $\hat{\mathbf{C}}_{i,m}$ based on object-centric features fused with a global latent embedding $\mathbf{z}_{obj}$. Using these canonical coordinates and the observed keypoints, the pose head estimates $(\mathbf{R},\mathbf{t},s)$, while relative $\mathrm{SE}(3)$ is computed after inference through a weighted Umeyama solver~\cite{umeyama1991least}. To recover real-world scale, the model supports a relative-scale head that infers scale from RGB features and the object latent $z_{obj}$, and an absolute-scale head that uses sensor-depth point clouds to predict translation and object size in the camera frame if available. This unified design provides camera parameters, depth maps, point maps, canonical coordinates, and both absolute and relative pose within one coherent framework. 

For the geometric supervision of camera extrinsics, point maps, and depth maps in normalized space, we follow \cite{wang2025vggt}. The depth loss follows the aleatoric-uncertainty formulation and uses the predicted uncertainty map $\Sigma_i^{D}$ to weight both the depth residual and the spatial gradient residual. The loss is 
\resizebox{\linewidth}{!}{$
\begin{aligned}
\mathcal{L}_{\text{depth}} =
\sum_{i=1}^{N} \Big(
 \big\| \Sigma_i^{D} \odot (\hat D_i - D_i) \big\|
 + \big\| \Sigma_i^{D} \odot (\nabla \hat D_i - \nabla D_i) \big\|
 - \alpha \log \Sigma_i^{D}
\Big)
\end{aligned}
$}
where $\odot$ denotes channel-broadcast element-wise multiplication.
The point-map loss uses the same structure, but with the point-map uncertainty 
$\Sigma_i^{P}$:
\resizebox{\linewidth}{!}{$
\begin{aligned}
\mathcal{L}_{\text{point}}
=
\sum_{i=1}^{N}
\Big(
\lVert \Sigma_i^{P} \odot (\hat{P}_i - P_i) \rVert
+
\lVert \Sigma_i^{P} \odot (\nabla \hat{P}_i - \nabla P_i) \rVert
-
\alpha \log \Sigma_i^{P}
\Big).
\end{aligned}
$}

\subsection{Keypoint-level Multi-view Feature Fusion}
Direct dense pixel-level NOCS regression with attention~\cite{vaswani2017attention} is expensive and sensitive to noise. Following keypoint-based formulations~\cite{lin2024instance,Li_2025_CVPR}, we represent each object by a compact set of $M$ latent keypoints that attend to joint visual and geometric evidence. Given a foreground RGB-D crop, we sample $N$ pixels and lift them to camera-space points $\mathbf{X}_k$ with associated colors and normals. A transformer backbone extracts image features $\mathbf{f}^{\text{rgb}}_{k}$ at the sampled tokens, while a 3D backbone~\cite{wu2025sonata} processes $(\mathbf{X}_k,\mathbf{I}_k,\mathbf{n}_k)$ to produce geometric features $\mathbf{f}^{\text{geo}}_{k}$. Concatenation yields local descriptors $\mathbf{f}_k = [\mathbf{f}^{\text{rgb}}_{k} \Vert \mathbf{f}^{\text{geo}}_{k}]$. A learnable query performs cross-attention using cosine similarity, producing soft heatmaps $\mathbf{H}_{m,k}$. Each keypoint is $\mathbf{X}^{\text{obj}}_m = \mathbf{H}_{m,k}\mathbf{X}^{\text{obj}}_k$, keypoint feature is extracted as $\mathbf{F}^{\text{obj}}_m = \mathbf{H}_{m,k}[\mathbf{f}^{\text{rgb}}_k \Vert \mathbf{f}^{\text{geo}}_k]$. A visual-geometric fusion block~\cite{lin2024instance} then refines $\mathbf{F}^{\text{obj}}_{m}$ by performing KNN grouping in 3D around $\mathbf{X}^{\text{obj}}_{m}$, encoding relative offsets and absolute coordinates, and applying cosine-similarity attention over the local neighborhoods. This aggregation yields enhanced keypoint descriptors $\tilde{\mathbf{F}}^{\text{obj}}_{m}$ that carry both local geometric context and global object cues. 

We further aggregate keypoint features across frames using a cross-frame attention module. Concretely, we augment keypoint descriptors with frame-wise sinusoidal positional encodings and process them with a transformer encoder, allowing keypoints from different views to exchange information at the feature level. In parallel, we pool the object latent embedding across the input views and share it back to each frame. This design enables multi-view geometric reasoning, enforces cross-view consistency, and improves absolute pose estimation by reducing single-view ambiguity. To ensure geometric consistency, we constrain $\mathbf{X}^{\text{obj}}_{m}$ to lie on the object surface using a Chamfer distance loss $\mathcal{L}_{\text{cd}}$.
\begin{equation}
\mathcal{L}_{\text{cd}} = 
\frac{1}{|\mathbf{X}^{\text{obj}}_{\text{m}}|} 
\sum_{x \in \mathbf{X}^{\text{obj}}_{\text{m}}} 
\min_{y \in {\mathbf{X}^{\text{obj}}_k}^\star} \|x - y\|_2^2.
\label{equ:cd}
\end{equation}
To prevent keypoints from collapsing into a small region, we add a diversity regularization that balances surface adherence and spatial diversity
\begin{equation}
\mathcal{L}_{\text{div}} = 
\frac{1}{M(M-1)} \sum_{x \neq y \in \mathbf{X}^{\text{obj}}_{\text{m}}} 
\max \big( 0, \tau_2 - \|x - y\|_2 \big)^2,
\label{equ:div}
\end{equation}
where $\tau_2$ controls the minimum separation between keypoints. To encourage keypoints to be representative of the depth-lifted point cloud, we employ a lightweight reconstruction head that takes keypoint positions and features $\tilde{\mathbf{F}}^{\text{obj}}_{m}$ as input, applies positional encoding, and decodes per-point displacement deltas to recover the object geometry. The reconstruction loss is a one-sided Chamfer distance between the reconstructed point cloud $\hat{\mathbf{X}}^{\text{obj}}$ and the observed camera-space points $\mathbf{X}^{\text{obj}}_k$:
\begin{equation}
\mathcal{L}_{\text{rec}}=\frac{1}{\left|\hat{\mathbf{X}}^{\text{obj}}\right|} 
\sum_{x \in \hat{\mathbf{X}}^{\text{obj}}} 
\min_{y \in {\mathbf{X}^{\text{obj}}_k}^{\star}}\left\|x-y\right\|_2.
\label{equ:loss_rec}
\end{equation}
The keypoint regularization is $\mathcal{L}_{\text{kpt}} = \mathcal{L}_{\text{cd}} + \mathcal{L}_{\text{div}} + \mathcal{L}_{\text{rec}}$.

\subsection{Canonical Correspondences \& Absolute Poses}
Given latent keypoint features, the NOCS head predicts canonical coordinates for each keypoint as $\hat{\mathbf{C}}_{m} = \mathrm{NOCS}(\tilde{\mathbf{F}}^{\text{obj}}_{m},\mathbf{z}_{\text{obj}})$. NOCS regression is addtionally conditioned on FiLM based affine transformation with parameters $(\gamma, \beta)$ to intermediate features, so that canonicalization is conditioned on the object code $\mathbf{z}_{\text{obj}}$ but remains category-agnostic. Absolute pose is estimated by an MLP-based pose and size head that takes $(\hat{\mathbf{C}}_{m}, \mathbf{X}^{\text{obj}}_{m})$ and the corresponding keypoint features $\tilde{\mathbf{F}}^{\text{obj}}_{m}$ as input. Rotation is represented in 6D~\cite{Zhou_2019_CVPR} and mapped to $\mathrm{SO}(3)$ via orthogonalization, while translation is predicted as a residual with respect to the point-cloud center following~\cite{lin2023vi,DualPoseNet}. An isotropic scale $\hat{s}$ is obtained from a per-axis size vector $\hat{\mathbf{s}}$ by averaging its magnitudes. The resulting homogeneous transform $\hat{\mathbf{S}} = [\hat{s}\hat{\mathbf{R}}, \hat{\mathbf{t}}]$ maps canonical coordinates to the camera frame. To supervise the normalized scale prediction, we use
\begin{equation}
\mathcal{L}_{\text{pose}} = \|\mathbf{R}_{\text{gt}} - \mathbf{R}\|_{\rm F} + \|\mathbf{t}_{\text{gt}} - \mathbf{t}\|_2 + \|\mathbf{s}_{\text{gt}} - \mathbf{s}\|_2.
\end{equation}
For $\mathcal{L}_{\text{nocs}}$, we use the Smooth $L_1$ loss with
\begin{equation}
\mathcal{L}_{\text{nocs}} = \| \mathbf{C}_{\text{m}}^{\text{gt}} - \mathbf{C}_{\text{m}}^{\text{nocs}} \|_{\text{SL1}}
\end{equation}

\subsection{Relative Poses from Depth and Point Map}
\label{subsec:relative}
We estimate the metric relative pose by aligning two independently predicted 3D structures with a robust, weighted Procrustes/Umeyama procedure. Let $(\mathrm{a},\mathrm{q})$ denote the anchor and query frames. Our model predicts two-view point maps $\mathbf{P}^{\mathrm{a}}$ and $\mathbf{P}^{\mathrm{q}}$ in the anchor coordinate system, while depth and intrinsics yield camera-space point clouds $\mathbf{X}^{\mathrm{a}}_{\text{cam}}$ and $\mathbf{X}^{\mathrm{q}}_{\text{cam}}$. We proceed in two steps:
\begin{enumerate}
  \item \textbf{Anchor calibration (Sim(3)).} We compute a weighted Umeyama similarity transform $\mathbf{S}_{\mathrm{a}}\in\mathrm{Sim}(3)$ that aligns the predicted anchor point map to the depth-derived anchor camera points,
 \begin{equation}
  \mathbf{S}_{\mathrm{a}} \;=\; \operatorname*{argmin}_{\mathbf{S}\in\mathrm{Sim}(3)} \sum_{n} w_n \,\big\lVert \mathbf{S}\,\mathbf{P}^{\mathrm{a}}_n - \mathbf{X}^{\mathrm{a}}_{\text{cam},n} \big\rVert_2^2
\end{equation}
  where weights $w_n$ come from point map confidences. We then apply $\mathbf{S}_{\mathrm{a}}$ to both $\mathbf{P}^{\mathrm{a}}$ and $\mathbf{P}^{\mathrm{q}}$, removing global scale ambiguity due to projective geometry ambiguity for uncalibrated camera.
  \item \textbf{Query alignment (SE(3)).} We then align the calibrated query point map $\mathbf{S}_{\mathrm{a}}\mathbf{P}^{\mathrm{q}}$ to the query camera-space points with a \emph{fixed-scale} Umeyama (i.e., $\mathrm{SE}(3)$),
\begin{equation}
  \mathbf{T}^{\mathrm{a}\rightarrow \mathrm{q}} \;=\; \operatorname*{argmin}_{\mathbf{T}\in\mathrm{SE}(3)} 
  \sum_{n} \tilde{w}_n \,\big\lVert 
  \mathbf{T}\,(\mathbf{S}_{\mathrm{a}}\mathbf{P}^{\mathrm{q}}_n) 
  - \mathbf{X}^{\mathrm{q}}_{\text{cam},n} 
  \big\rVert_2^2.
\end{equation}
\end{enumerate}
The resulting $\mathbf{T}^{\mathrm{a}\rightarrow \mathrm{q}}$ is the relative pose from anchor to query. Given an absolute pose $\mathbf{T}^{\mathrm{a}}$ for the anchor, the query absolute pose is recovered as
$\mathbf{T}^{\mathrm{q}}=\mathbf{T}^{\mathrm{a}\rightarrow \mathrm{q}}\mathbf{T}^{\mathrm{a}}$. This two-step formulation (Sim(3) followed by fixed-scale $\mathrm{SE}(3)$) is robust to monocular scale ambiguity.

\subsection{Contrastive Learning for Object Latent}
We train the object latent embedding $\mathbf{z}_{obj}$ using a supervised InfoNCE objective to enhance the alignment of samples belonging to the same instance or semantic category. Let $\{z_i\}_{i=1}^{N}$ denote the normalized latent features $\mathbf{z}_{obj}$ in a batch, the pairwise similarity is defined as $s_{ij} = \frac{z_i^{\top} z_j}{\tau}$, where $\tau$ is a fixed temperature. A binary mask $P \in \{0,1\}^{N \times N}$ specifies the positive pairs. The diagonal entries of $P$ are zero. The denominator for each anchor $i$ includes all non-diagonal terms $d_i = \log \left( \sum_{j \neq i} \exp(s_{ij}) \right)$. Let $w$ be a weight matrix aligned with $P$, and let $\epsilon$ be a small constant used for stability. The numerator in the log-sum mode is 
$n_i=\log(\sum_{j:P_{ij}=1}\exp(s_{ij}+\log(w_{ij}+\epsilon)))$.
The final supervised InfoNCE loss is computed as
\begin{equation}
    \mathcal{L}_{\text{InfoNCE}}
    = -\frac{1}{|\mathcal{V}|}
    \sum_{i \in \mathcal{V}} \left( n_i - d_i \right),
\end{equation}
where $\mathcal{V} = \{ i \;|\; \sum_j P_{ij} > 0 \}$ denotes the anchors that contain at least one positive match. Anchors without any positive pairs are excluded from the average.

During distributed training, we gather all latent features across devices before computing the loss, which enlarges the set of negatives and keeps the loss consistent across ranks. The positive mask and weight matrix are expanded to match the aggregated latent features. This produces a stable latent space where samples of the same object or category become closer while unrelated samples remain separated.

\subsection{Metric Scale Recovery \& Camera Parameters}
\label{subsec:scale}
Monocular predictions are scale-ambiguous. With sensor depth $D^{\text{sens}}$, we sample $K$ points at $\mathcal{I}_i$ and form a centered point cloud $\{\tilde{\mathbf{x}}_k\}$. An absolute-scale head encodes $\{\tilde{\mathbf{x}}_k\}$ with PointNet++~\cite{qi2017pointnetplusplus}, fuses the features with a projected $\mathbf{z}_{\text{obj}}$, and predicts \emph{absolute} camera-frame translation $\hat{\mathbf{t}}^{\text{abs}}$ and size $\hat{\mathbf{s}}^{\text{abs}}$ and supervise the translation ans size in absolute scale using L1 loss. When depth is absent, a lightweight head regresses a per-frame log-scale from global visual descriptors fused with $\mathbf{z}_{\text{obj}}$ and outputs a confidence. This branch is auxiliary during training and can be used at inference to provide a scale prior in pure RGB mode. We supervise these with an L1 loss
{\footnotesize
$\mathcal{L}_{\text{scale}}
=
\lVert \hat{\mathbf{t}}^{\text{abs}} - \mathbf{t}^{\text{abs}} \rVert_1
+
\lVert \hat{\mathbf{s}}^{\text{abs}} - \mathbf{s}^{\text{abs}} \rVert_1
+
\lVert \log \hat{s} - \log s^* \rVert_1$.
}

The camera head predicts a compact pose encoding per frame that decodes to intrinsics and supports differentiable back-projection. Training uses an $\ell_1$ loss on focal lengths and on the extrinsic represented by a quaternion, we supervise the pose encoding through a Huber distance between the predicted camera parameters $\hat{\mathbf{g}}_i$ and the ground truth $\mathbf{g}_i$:
$\mathcal{L}_{\text{cam}}=\sum_{i=1}^{N}\lVert \hat{\mathbf{g}}_i-\mathbf{g}_i\rVert_{\epsilon}$.

\subsection{Overall Loss Function} \label{subsec:pse}
\label{subsec:loss}
We supervise multi-head predictions with a combination of losses:
\begin{equation}
\begin{aligned}
\mathcal{L}
= &\,\mathcal{L}_{\text{cam}}
+ \mathcal{L}_{\text{depth}}
+ \mathcal{L}_{\text{point}}
+ \mathcal{L}_{\text{nocs}}
+ \mathcal{L}_{\text{kpt}} \\
&+ \mathcal{L}_{\text{pose}}
+ \mathcal{L}_{\text{scale}}
+ \mathcal{L}_{\text{InfoNCE}}
\end{aligned}
\end{equation}
We provide detailed settings in the supplementary material.
\section{Experiment} \label{sec:experiment}

\begin{figure*}[t]
    \centering
    \includegraphics[width=0.95\textwidth]{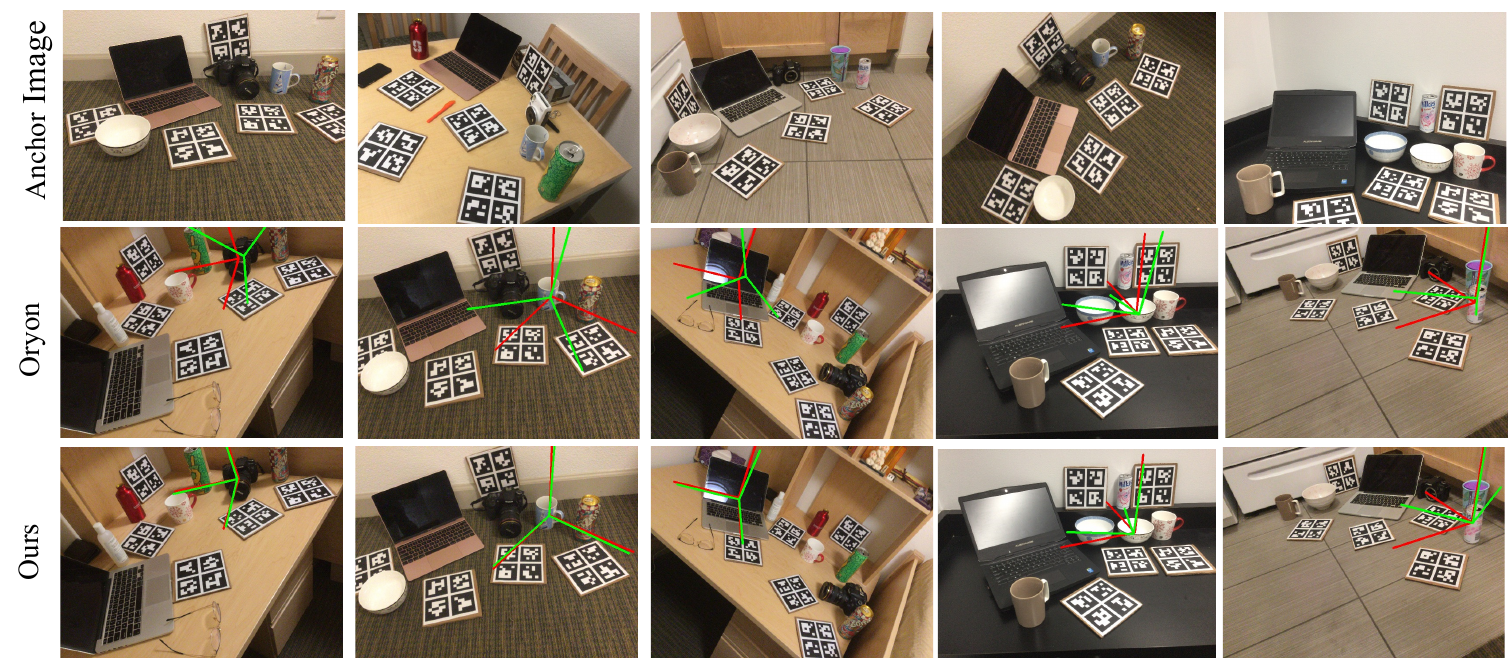}
    \caption{Qualitative result in relative pose estimation. We compare different object instances across different scenes with Oryon \cite{corsetti2024open}. Visualization shows that our OPT-pose can estimate the relative object poses across different objects and scenes.}
    \label{fig:relpose}
    \vspace{-0.2cm}
\end{figure*}

\begin{figure*}[t]
    \centering
    \includegraphics[width=0.95\textwidth]{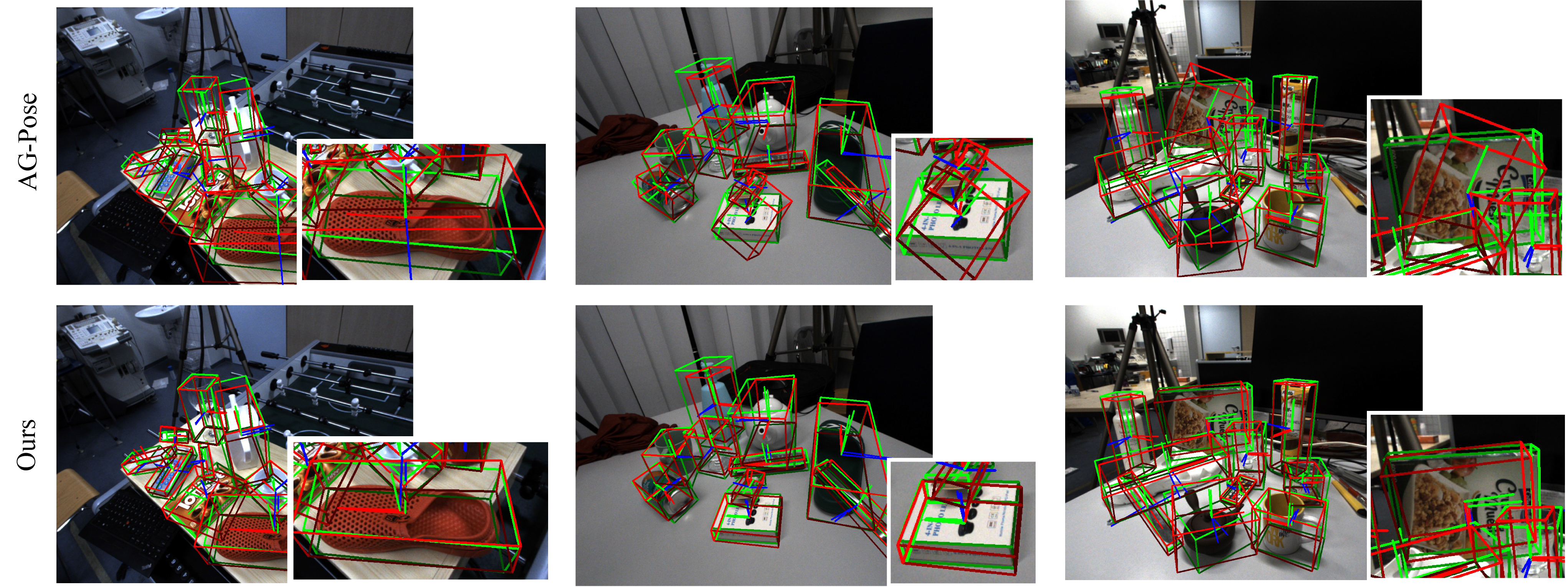}
    \vspace{-0.2cm}
    \caption{Qualitative result in absolute pose estimation with RGB-D input. We showcase some difficult instances in comparison with AG-Pose\cite{lin2024instance}. We zoom in on the difficult object categories, the shoe and the box, for better visualization.}
    \label{fig:abs_pose}
    \vspace{-0.3cm}
\end{figure*}

\begin{table}
\centering
\scriptsize
\setlength{\tabcolsep}{3pt}
\caption{Category-level pose estimation (RGB) on REAL275 using scale-agnostic metrics.}
\vspace{-0.2cm}
\label{tab:main_real}
\begin{tabular}{c c | c c c c c c c c}
    \toprule
        Method  & \textbf{NIoU75} & \textbf{5°0.2d}  & \textbf{5°0.5d} & \textbf{10°0.2d} & \textbf{10°0.5d} & \textbf{0.2d} & \textbf{5°} & \textbf{10°} \\ \hline
        MSOS ~\cite{msos} & \hphantom{0}0.7 & - & - &\hphantom{0}3.3 & 15.3 & 10.6 & - & 17.0 \\ 
        OLD-Net ~\cite{oldnet}& \hphantom{0}0.4 & \hphantom{0}0.9 & \hphantom{0}3.0 & \hphantom{0}5.0 & 16.0 & 12.4 & \hphantom{0}4.2 & 20.9  \\
        DMSR ~\cite{dmsr} & \hphantom{0}9.5 & 15.1 & 23.7  & 25.6 & 45.2 & 35.0 & 27.4 & 52.0\\ 
        {LaPose} ~\cite{zhang2024lapose}& 15.8 & 15.7 &  21.3& 37.4 & 57.4 & 46.9 & 23.4 & 60.7\\ 
        {GIVE-Pose} ~\cite{huang_2025_GIVEPose} & \textbf{20.8} & - &  - & \textbf{44.6} & {64.8} & {46.9} & - & 67.8\\ 
        {UniDet} ~\cite{fischer2025unified} & \underline{19.2} & \textbf{25.1}& {31.8}  & {43.7} &\underline{66.1} & \underline{53.5}  & \underline{32.1} & \underline{68.8}\\
        \hline
        {\textbf{\ours{}}}  & {11.1} & \underline{24.3}& \textbf{52.2}  & \underline{44.1} &\textbf{82.3} & \textbf{56.4} &  \textbf{52.3} & \textbf{82.5}\\
        \bottomrule
    \end{tabular}
\end{table}

\begin{table*}[t]
\centering
\footnotesize
\caption{\textbf{Category-level absolute pose estimation (RGB-D) on HouseCat6D.} 
Models are trained on respective training sets and evaluated on the test set. 
Prior columns indicate: \textbf{Category} (predefined category input), 
\textbf{Shape} (shape prior), \textbf{Semantic} (semantic prior), and 
\textbf{Calibration} (camera intrinsics). 
\textbf{MV} indicates whether the method supports multi-view pose reasoning. 
\textbf{Bold} and \underline{underlined} denote the best and second-best results, respectively. 
We additionally report \ours{} with multi-view enhanced inference ($\mathbf{S}{=}2,3,4$), 
which is only applicable to our method and demonstrates its ability to leverage 
cross-view geometric cues for absolute pose estimation.}
\vspace{-0.2cm}
\label{tab:merged_results}
\resizebox{\textwidth}{!}{
\begin{tabular}{l|l|c c c c c|c c c c|c c}
\toprule
\textbf{Dataset} & \textbf{Method} 
& \textbf{Category} & \textbf{Shape} & \textbf{Semantic} & \textbf{Calibration} 
& \textbf{MV} 
& \textbf{5$^\circ$2cm} & \textbf{5$^\circ$5cm} & \textbf{10$^\circ$2cm} & \textbf{10$^\circ$5cm} 
& \textbf{IoU50} & \textbf{IoU75} \\
\midrule
\multirow{15}{*}{HouseCat6D} & \multicolumn{12}{c}{\textbf{Single-view Inference (standard setting)}} \\
\cmidrule(lr){2-13}
& DPDN~\cite{lin2022category} & \checkmark & \checkmark &  & \checkmark &  & 6.4 & 6.9 & 22.2 & 25.8 & 56.2 & 26.0 \\
& VI-Net~\cite{lin2023vi} & \checkmark &  &  & \checkmark &  & 8.4 & 10.3 & 20.5 & 29.1 & 56.4 & 20.4 \\
& SecondPose~\cite{chen2024secondpose} & \checkmark &  & \checkmark & \checkmark &  & 11.0 & 13.4 & 25.3 & 35.7 & 66.1 & 24.9 \\
& AG-Pose~\cite{lin2024instance} & \checkmark &  &  & \checkmark &  & 11.5 & 12.0 & 32.7 & 35.8 & 66.0 & 45.0 \\
& Sphere-Pose~\cite{iclr2025spherepose} & \checkmark &  &  & \checkmark &  & 19.3 & 25.9 & 40.9 & 55.3 & 72.2 & - \\
& Spot-Pose~\cite{cvpr2025spotpose} & \checkmark &  &  & \checkmark &  & 23.8 & 24.5 & 52.3 & 54.8 & 77.0 & - \\
& GCE-Pose~\cite{Li_2025_CVPR} & \checkmark & \checkmark & \checkmark & \checkmark &  & 24.8 & 25.7 & \textbf{55.4} & 58.4 & \textbf{79.2} & \textbf{60.6} \\
& \textbf{\ours{}} ($\mathbf{S}{=}1$) &  &  &  &  & \checkmark & \underline{28.0} & \underline{31.9} & 53.0 & \underline{60.3} & 78.1 & 40.7 \\
& \textbf{OPT-Pose} ($\mathbf{K}_{\mathrm{gt}}$, $\mathbf{S}{=}1$) &  &  &  & \checkmark & \checkmark & \textbf{29.3} & \textbf{32.1} & \underline{55.1} & \textbf{60.4} & \underline{79.0} & \underline{52.2} \\
\cmidrule(lr){2-13}
& \multicolumn{12}{c}{\textbf{Multi-view Enhanced Inference (ours only)}} \\
\cmidrule(lr){2-13}
& \textbf{OPT-Pose} ($\mathbf{K}_{\mathrm{gt}}$, $\mathbf{S}{=}2$) &  &  &  & \checkmark & \checkmark & {34.1} & {37.0} & {55.8} & {61.2} & {79.6} & 54.3 \\
& \textbf{OPT-Pose} ($\mathbf{K}_{\mathrm{gt}}$, $\mathbf{S}{=}3$) &  &  &  & \checkmark & \checkmark & {36.1} & {38.9} & {56.1} & {61.7} & { \textbf{79.7}} & {54.8} \\
& \textbf{OPT-Pose} ($\mathbf{K}_{\mathrm{gt}}$, $\mathbf{S}{=}4$) &  &  &  & \checkmark & \checkmark & {\textbf{36.9}} & {\textbf{39.7}} & {\textbf{56.4}} & {\textbf{62.1}} & { \underline{79.6}} & {\textbf{54.9}} \\
\bottomrule
\end{tabular}
}
\label{tab:abs_housecat}
\vspace{-0.2cm}
\end{table*}

\begin{table}[t]
\centering
\scriptsize
\caption{\textbf{Results on Omni6DPose (ICCV 2025 WLCOP Challenge}~\cite{wclop2025challenge}). 
Our model is trained solely on Omni6DPose, whereas GenPose++~\cite{zhang2023genpose} and CPPF++~\cite{you2024cppf++} further leverage categorical augmentation from PACE~\cite{you2023pace}.}
\vspace{-0.3cm}
\setlength{\tabcolsep}{2.5pt}
\begin{tabular}{l|l|ccc|cccc}
\toprule
Dataset & Method 
& \multicolumn{3}{c|}{AUC $\uparrow$} 
& \multicolumn{4}{c}{VUS $\uparrow$} \\
\cline{3-9}
 & 
& IoU25 & IoU50 & IoU75 
& 5°2cm & 5°5cm & 10°2cm & 10°5cm \\
\midrule
\multirow{3}{*}{ROPE}
& GenPose++~\cite{zhang2024omni6dpose} & \underline{39.64} & \underline{18.68} & \underline{1.28} & 8.02 & 11.90 & 16.58 & 24.87 \\
& CPPF++~\cite{you2024cppf++}    & 38.19 & 17.18 & 1.07 & \textbf{9.43} & \textbf{14.20} & \textbf{18.28} & \textbf{27.66} \\
& \textbf{OPT-Pose ($\mathbf{K}_{\mathrm{gt}}$)} & \textbf{40.06} & \textbf{21.03} & \textbf{2.73} & \underline{9.08} & \underline{13.14} & \underline{17.62} & \underline{25.46} \\
\bottomrule
\end{tabular}
\label{tab:omni_pose_results}
\end{table}

\subsection{Implementation Details}
We follow the image configuration from \cite{wang2025vggt}. Images are resized to $518\times518$ with a patch size $14$. For each frame we sample $K{=}1024$ pixels to obtain indices $\mathcal{I}_i$ used by all heads. We predict $M{=}128$ keypoints. We set the softmax temperatures in keypoint attention and contrastive InfoNCE to $1.0$, use a repulsion threshold $0.02$ in the keypoint diversity loss, pose weight of $\epsilon{=}10^{-3}$ for absolute pose, Smooth-$\ell_1$ with threshold $0.1$ for sparse NOCS and relative-scale supervision (with $\beta{=}0.1$), and $k_{\!n}\in\{8,16,32\}$ neighbors per stage in the fusion block. The multiview transformer encoder is frozen, leaving the attention to fine-tune the features for canonicalization. The model is optimized with AdamW and parameter groups: NOCS/pose/fusion modules and projectors use a base learning rate of $5\times10^{-4}$ with weight decay of $1\times10^{-2}$; the global optimizer uses a learning rate of $5\times10^{-7}$ with weight decay of $0.05$. We apply 5\% linear warm-up then cosine decay, mixed precision, and per-module gradient clipping. We provide more parameter details in supplementary material.

\subsection{Benchmarks and Protocols}
We evaluate on three tasks to demonstrate unified categorical absolute and unseen-object relative pose in a single framework, with flexible RGB-(D) input and intrinsics:
\begin{itemize}
    \item \textbf{Category-level absolute pose (RGB-D):} \textbf{HouseCat6D} \cite{jung2024housecat6d}. We report metrics $(5^\circ,2\mathrm{cm})$, $(5^\circ,5\mathrm{cm})$, $(10^\circ,2\mathrm{cm})$, $(10^\circ,5\mathrm{cm})$ thresholds and 3D IoU. For \textbf{ROPE}~\cite{zhang2024omni6dpose}, we report: \textbf{VUS} (Volume Under Surface) with rotation thresholds from $1^\circ$ to $15^\circ$ and translation thresholds from $1\,\mathrm{cm}$ to $5\,\mathrm{cm}$, and \textbf{AUC} (Area Under Curve), which evaluates Intersection over Union (IoU) of 3D bounding boxes over IoU thresholds from $0.25$ to $0.95$.
  \item \textbf{Category-level absolute pose (RGB; scale-agnostic on REAL275):} Following prior work, we report normalized IoU ($NIoU$) and distance thresholds on REAL275 in RGB-only settings for scale-agnostic pose estimation.
  \item \textbf{Unseen-object relative pose (RGB-D):} \textbf{NOCS-REAL}, \textbf{Toyota-Light (TOYL)}. Trained on SOPE~\cite{zhang2024omni6dpose}, these benchmarks test SE(3) alignment across views for unseen objects. We evaluate using ADD(-S), AR, MSSD, MSPD, and VSD metrics; relative SE(3) is estimated post-hoc via weighted Umeyama alignment of depth and point-map structures (Sec.~\ref{subsec:relative}).
\end{itemize}

\subsection{Comparison with the State of the Art}
\paragraph{Category-level absolute pose (RGB; scale-agnostic REAL275).}
As shown in Tab.~\ref{tab:main_real}, our method substantially outperforms prior work on REAL275 in scale-agnostic evaluation. We achieve great performance on all reported metrics except $NIoU_{75}$, where we are comparable to UniDet\cite{fischer2025unified} and surpass GIVEPose~\cite{huang_2025_GIVEPose}, DMSR\cite{dmsr}, LaPose\cite{zhang2024lapose} by in-degree/normalized distance thresholds. Note that these methods explicitly use the predefined calculated size and inference with category priors.

\paragraph{Category-level absolute pose (RGB-D).} Leveraging measured depth, \ours{} recovers metric scale via the absolute-scale head and achieves strong performance on HouseCat6D~\cite{jung2024housecat6d} (See \cref{fig:abs_pose}). In Tab.~\ref{tab:merged_results}, we obtain the best results under strict thresholds ($5^\circ$2cm, $5^\circ$5cm) and competitive accuracy at looser thresholds and IoU compared to GCE-Pose\cite{Li_2025_CVPR}, while using no category, shape, semantic, or calibration priors. In addition, we evaluate a multi-view enhanced inference mode using $S{=}2,3,4$ frames without retraining. As shown in Tab.~\ref{tab:merged_results}, multi-view inference consistently improves absolute pose accuracy across all metrics. This supports our central design: relative geometric reasoning across views provides additional constraints that reduce ambiguity in single-view predictions, leading to more stable and accurate absolute pose estimation. To validate the practical utility of our unified framework for real-world downstream tasks, we evaluate \ours{} on the large-vocabulary Omni6DPose benchmark, designed for robotic manipulation. As shown in Tab.~\ref{tab:omni_pose_results}, \ours{} surpasses recent state-of-the-art methods like GenPose++~\cite{zhang2024omni6dpose} and CPPF++~\cite{you2024cppf++} in strict IoU metrics and achieves highly competitive accuracy at 5°/10° thresholds while using less prior and performing inference with a category-agnostic setting. This demonstrates that our category-agnostic canonicalization effectively scales to large-vocabulary scenarios, which are critical for robotics.

\paragraph{Unseen-object relative pose.} For unseen objects pose estimation, \ours{} aligns depth and point-map branches with a weighted Umeyama to yield robust SE(3) estimation across frames. As shown in Tab.~\ref{tab:model_free_pose}, our method outperforms all existing methods by large margins across ADD(-S), AR, MSSD, MSPD, and VSD, especially on NOCS-REAL (\cref{fig:relpose}), demonstrating that task factorization within a single model enables strong performance on both canonical-space (absolute) and camera-space (relative) reasoning. The slight performance gain in TOY-L is due to our method being more sensitive to illumination changes, and to geometric alignment being less accurate under these conditions.

\subsection{Ablation Studies}
We analyze key design choices on HouseCat6D to validate our unified model-free formulation with task factorization.
\begin{itemize}
  \item \textbf{Object latent embedding.} Removing contrastive learning harms canonical correspondence stability and reduces HouseCat6D accuracy (Tab.~\ref{tab:abs_housecat}), showing that object latent learning is key for category-agnostic behavior without predefined category names during inference.
  \item \textbf{Pointmap head.} Removing the pointmap head causes negligible change in absolute pose accuracy, which is expected: the pointmap branch serves relative SE(3) estimation, not absolute pose. This confirms that the two geometric pathways are decoupled by design — the absolute pose pathway (Depth + NOCS) is not degraded by the addition of the relative pose pathway, demonstrating that task factorization enables both capabilities within a single model at no accuracy cost to either task.
  \item \textbf{Metric head} Replacing the absolute metric head with a test-time Umeyama algorithm for scale calculation weakens pose accuracy under metric evaluation, indicating that learning absolute translation/size from depth is necessary.
  \item \textbf{Keypoint extraction} Replacing the keypoint extraction with direct pixel-level DPT, with sensor depth aligned, significantly drops the performance, showcasing that the dense prediction may drop the performance due to noise(prediction, depth).
  \item \textbf{Multi-view reasoning (~\cref{tab:abs_housecat})} Increasing the number of input views at inference time consistently improves absolute pose estimation without retraining, indicating that relative geometric reasoning provides complementary constraints beyond single-view predictions.
\end{itemize}


\begin{table}[t]
\centering
\scriptsize
\caption{\textbf{Relative Object Pose Estimation Results.} AUC of ADD(-S), AR, MSSD, MSPD, and VSD on Real275 and Toyota-Light.}
\label{tab:model_free_pose}
\setlength{\tabcolsep}{3pt}
\begin{tabular}{l|l|ccccc}
\toprule
\textbf{Dataset} & \textbf{Method} & \textbf{ADD(-S)} & \textbf{AR} & \textbf{MSSD} & \textbf{MSPD} & \textbf{VSD} \\
\midrule

\multirow{6}{*}{REAL275}
& SIFT~\cite{lowe1999object} & 16.4 & 34.1 & 37.9 & 48.0 & 16.5 \\
& Obj. Mat.~\cite{gumeli2023objectmatch} & 13.4 & 26.0 & 31.7 & 30.8 & 15.5 \\
& Oryon~\cite{corsetti2024open} & 34.9 & 46.5 & 50.9 & 56.7 & \underline{32.1} \\
& Any6D~\cite{lee2025any6d} & \underline{53.5} & 51.0 & \underline{56.5} & \underline{65.3} & 31.1 \\
& One2Any~\cite{liu2025one2any} & 41.0 & \underline{54.9} & - & - & - \\
& \textbf{\ours{}} & \textbf{94.2} & \textbf{84.2} & \textbf{89.3} & \textbf{91.3} & \textbf{71.9} \\

\midrule

\multirow{5}{*}{Toyota-Light}
& SIFT~\cite{lowe1999object} & 14.1 & 30.3 & 39.6 & 44.1 & 7.3 \\
& Obj. Mat.~\cite{gumeli2023objectmatch} & 5.4 & 9.8 & 13.0 & 14.0 & 2.4 \\
& Oryon~\cite{corsetti2024open} & 22.9 & 34.1 & 42.9 & 45.5 & 13.9 \\
& Any6D~\cite{lee2025any6d} & {32.2} & \underline{43.3} & \underline{55.8} & \underline{58.4} & \underline{15.8} \\
& One2Any~\cite{liu2025one2any} & \underline{34.6} & {42.0} & - & - & - \\
& \textbf{\ours{}} & \textbf{47.0} & \textbf{57.1} & \textbf{59.4} & \textbf{62.8} & \textbf{49.0} \\

\bottomrule
\end{tabular}
\end{table}

\begin{table}[t]
\centering
\scriptsize
\caption{Ablations for single-view category-level absolute pose on HouseCat6D (RGB-D). Metrics follow Sec.~\ref{sec:experiment}.}
\vspace{-0.2cm}
\label{tab:abl_abs_hc}
\begin{tabular}{l|cccc}
\toprule
{HouseCat6D (RGB-D)} & \textbf{5°2cm} & \textbf{5°5cm} & \textbf{10°2cm} & \textbf{10°5cm } \\
\midrule
Full pipeline & 28.0 & 31.9 & 53.0 & 60.3  \\
\; w/o pointhead & 27.9 & 30.6 & 54.5 & 58.8   \\
\; w/o object latent embedding & 24.5 & 26.7 & 48.9 & 52.6   \\
\; w/o Abs. Metric Head & 22.2 & 24.2 & 48.8 & 52.2  \\
\; w/o Keypoint Extration (DPT) & 20.0 & 22.9 & 47.0 & 54.8   \\
\bottomrule

\end{tabular}
\vspace{-0.2cm}
\end{table}

\section{Conclusions and Limitations} \label{sec:conclusion}
We presented Object Pose Transformer (\ours{}), the first unified model-free framework for task-factorized unseen object pose estimation. \ours{} bridges two previously fragmented paradigms: category-level absolute pose and unseen-object relative pose estimation. Our approach employs a single feed-forward model that predicts point maps, depth, NOCS, and camera parameters from RGB images. The complementary pairing of Depth+NOCS and Depth+Pointmap achieves canonical and relative pose reasoning. Evaluated on diverse datasets, OPT-Pose delivers state-of-the-art accuracy on both category-level absolute and unseen-object relative pose benchmarks within a unified architecture. Although \ours{} achieves strong performance on model-free tasks, the model shares limitations with other works in this domain: it requires object-centric crops, and performance degrades with large illumination changes. Moreover, model canonicalization can be ambiguous under symmetry or when the dataset convention changes. We believe this unified model-free framework with task factorization opens a promising direction for generic, category- and camera-agnostic object pose understanding, where future work may explore large-scale pretraining, multi-object reasoning, and tighter integration with robotic manipulation.

{
    \small
    \bibliographystyle{ieeenat_fullname}
    \bibliography{main}
}


\end{document}